%% file: MICCAI17-forGE.tex
\documentclass{llncs}
\usepackage{amssymb} %
\usepackage{amsmath} 
\usepackage{amsfonts}
\usepackage{mathrsfs}
\usepackage{gensymb}
\usepackage{graphicx} 
\usepackage{sidecap} 
\usepackage{tabularx}
\usepackage{floatrow}
\newfloatcommand{capbtabbox}{table}[][\FBwidth]
\usepackage{booktabs}
\usepackage{algpseudocode}
\usepackage{algorithm}
\usepackage{epstopdf}
\makeatletter
\def\BState{\State\hskip-\ALG@thistlm}
\makeatother
\usepackage{multirow}
\usepackage{url}

\newif\ifalgo
\algofalse

\begin{document}

\title{Quantification of Metabolites in Magnetic Resonance Spectroscopic Imaging using Machine Learning}
\titlerunning{Machine Learning for MRSI}

\author{Dhritiman Das$^{1, 3}$, Eduardo Coello$^{2, 3}$, Rolf F. Schulte$^{3}$, Bjoern H. Menze$^{1}$}

\authorrunning{Das et al.}

\institute
{%
$^1$Department of Computer Science, Technical University of Munich, Germany\\
$^2$Department of Physics, Technical University of Munich, Germany \\
$^3$GE Global Research Europe, Munich, Germany
}

\maketitle


\begin{abstract}
	Magnetic Resonance Spectroscopic Imaging (MRSI) is a clinical imaging modality for measuring tissue metabolite levels in-vivo. An accurate estimation of spectral parameters allows for better assessment of spectral quality and metabolite concentration levels. The current gold standard quantification method is the LCModel - a commercial fitting tool. However, this fails for spectra having poor signal-to-noise ratio (SNR) or a large number of artifacts. This paper introduces a framework based on random forest regression for accurate estimation of the output parameters of a model based analysis of MR spectroscopy data. The goal of our proposed framework is to learn the spectral features from a training set comprising of different variations of both simulated and in-vivo brain spectra and then use this learning for the subsequent metabolite quantification. Experiments involve training and testing on simulated and in-vivo human brain spectra. We estimate parameters such as concentration of metabolites and compare our results with that from the LCModel.
	
\end{abstract}


\section{Introduction} \label{sec:Intro}
Magnetic resonance spectroscopic imaging (MRSI) is an in-vivo clinical imaging modality which detects nuclear magnetic resonance signals produced by nuclei in living tissues. Quantification of this signal amplitude generates metabolic maps which show the concentration of metabolites in the sample being investigated. Accurate quantification of these metabolites is important for diagnosis of brain tumor and other in-vivo diseases. For this purpose, a common practice in the MRS community has been to use non-linear spectral fitting tools such as the LCModel \cite{Provencher1993}, TARQUIN \cite{wilson2011a}, AMARES \cite{VANHAMME199735} and ProFit \cite{Schulte2006} amongst which the LCModel is regarded as the gold standard fitting tool. In this study, we present an alternative to the non-linear model fitting using a machine learning approach.
 
\textbf{Non-linear model fitting.}
The LCModel software uses a linear combination of metabolite basis spectra set to model the spectral measurement in the frequency domain. It also uses smoothing splines to model the baseline signals and subsequently fits the parameters of the basis set using a non-linear optimisation. LCModel incorporates the prior knowledge of the data while modeling the fit and this ensures robustness in the model leading to estimation of the spectral parameters such as concentration of metabolites. Some of the drawbacks of this non-linear fitting model are: (1) Metabolite quantification can be time-consuming depending on the dataset size and requires a lot of manual parameter tuning. (2) The error in estimating parameters is lower if high SNR spectra are used since the non-linear voxel-wise fitting to noisy data leads to a high amount of local minima and subsequent inaccuracy in quantification. \cite{MRM:MRM21519}\cite{Kelm2012}.

 \textbf{Machine Learning.}
 Machine learning methods such as decision forests, random forests \cite{Breiman2001} are being extensively used in the medical imaging community for tasks such as parameter estimation, diseases diagnosis, segmentation, etc. In MRSI, machine learning tools have been used only for specific tasks such as classification of spectra \cite{MRM:MRM21519} and assessment of spectral quality \cite{MRM:MRM26618}. This opens up the possibility of using the recent advances in machine learning to predict MRSI data parameters while addressing the drawbacks of conventional fitting tools such as long computation time and poor performance for data with artifacts.
 
\textbf{Our Contribution.}
 In this work, we propose a simple yet effective method using random forest regression for multi-parameter estimation in MR Spectroscopic Imaging. We generate over 1 million simulated spectra training-set having concentration magnitudes, linewidth effects, baseline and lipid artifacts. We also use spectral data from 287 human subjects to create a physical training model to be used in the regression framework (Sect. \ref{Real}). In the following we present our method adapting random forest regression to MRSI (Sect. \ref{sec:Methods}) followed by experiments in the aforementioned dataset. Our proposed method is then validated quantitatively and qualitatively using: (1) synthetic brain spectra, (2) human in-vivo single voxel spectra  having the same image acquisition protocol as the physical training model and (3) independently acquired human in-vivo 2D MRS Images to perform a blind test on the physical and synthetic models. We present the results (Sect. \ref{sec: Results}) of our experiments followed by a summary and discussion (Sect. \ref{sec:Conclusion}) on the future work in this domain. This is the first application- to the best of our knowledge- of machine learning for determining MRS parameters which were otherwise determined using basis fitting tools.

\section{Methods} \label{sec:Methods}

\begin{figure}[ht!]
  \centering
 \includegraphics[width= \textwidth, height=3.8cm, keepaspectratio]{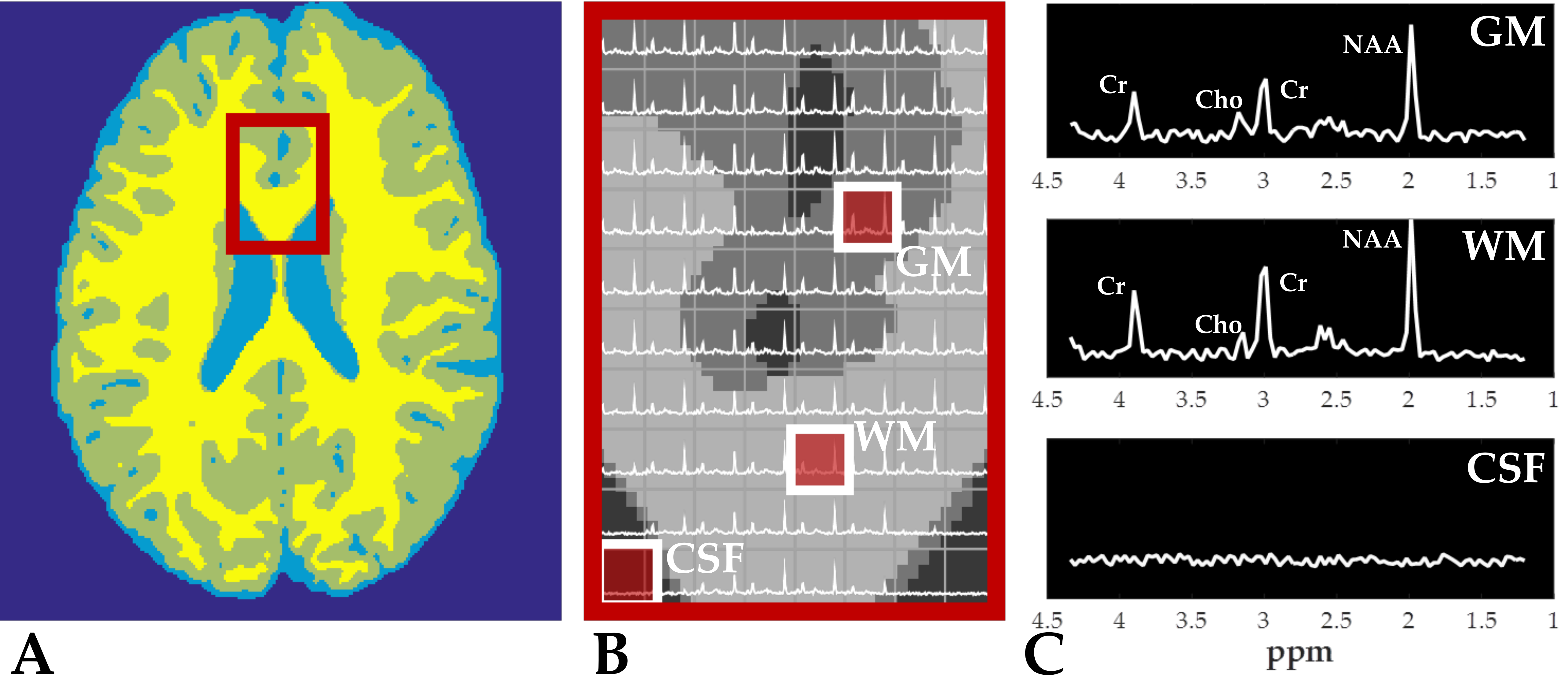}
 \caption{Example brain 2D MRSI dataset. \text(A) The simulated brain with the region of interest (red box). \text(B) Highlighted regions corresponding to GM, WM and CSF \text(c) Corresponding spectrum of GM, WM and CSF.\label{fig:BrainWeb}}
\end{figure}

\textbf{MR Spectroscopy.} 
Magnetic resonance spectroscopy, based on the concept of nuclear magnetic resonance (NMR), exploits the resonance frequency of a molecule, to obtain information about the concentration of a particular metabolite \cite{RobinA.deGraaf2013}. The time-domain complex signal of a nuclei is given by: 
\begin{equation}
\label{MRS}
S(t) = \int \mathrm{p}(\omega)\mathrm{exp}(-i\Phi)\mathrm{exp}(-t/T^{*}_{2})dw.
\end{equation}
The frequency-domain signal is given by $S(\omega)$, $T^{*}_{2}$ is the magnetization decay in the transverse plane due to magnetic field inhomogeneity and $\mathrm{p}(\omega)$ comprises of Lorentzian absorption and dispersion line-shapes function having the spectroscopic information about the sample.
$\Phi$ represents the phase, $(\omega t + \omega_{0})$, of the acquired signal where $\omega t$ is the time-varying phase change and $\omega_{0}$ is the initial phase.
Non-linear fitting tools facilitate the generation of metabolic maps to estimate concentration of metabolites such as N-acetyl-aspartate (NAA), Creatine (Cr) and Choline (Cho). An example of the spectra present in the brain has been shown in Fig. \ref{fig:BrainWeb}.

\textbf{Random Forest Regression.}
 Random Forests \cite{Breiman2001} have been shown to be effective in a wide range of classification and regression problems. These comprise of a set of binary trees wherein splits are created in each tree based on a random subsets of the feature variables on which the forests are subsequently trained. Piecewise linear regression is implemented by each tree over the input data and, after seeking for the best prediction at every node, data points are sent to the left or right branches based on feature selection by thresholding. This process continues till it reaches the end of the tree and subsequently the weighted average of the prediction from each tree is taken to give a single output estimate.
The randomness in the training process encourages the trees to give independent estimates which can be combined to achieve an accurate and robust result. 

For MRSI, we adapt the random forest approach to have a training dataset $D = (S_{i} (\omega), Y_{i})$ 
, $i\in [1, N]$, where $N$ is the total number of training spectra. 
$S_{i}(\omega)$ represents the training spectral data while $Y_{i}$ represents the corresponding multi-parameter training labels. For our model, we consider the concentrations of NAA, Cho and Cr for simulated data, while for the real data we additionally consider Myo-Inositol (mI) and Glutamate+Glutamine (Glx). Therefore, for a given spectra $S_{i}(\omega)$, $Y_{i} = [$ NAA$_{i}$, Cho$_{i}$, Cr$_{i}$, mI$_{i}$, Glx$_{i}$ $]$. 

Running the random forest regression on this produces a training model which can then be used to obtain parameter estimates $\hat{Y}_{j}$ of test spectra $S_{j}(\omega)$ having test labels $Y_{j}$, $j\in [1, M]$ where $M$ is the total number of test spectra.

\textbf{Error Calculation.}
For our experiments, given the estimate $\hat{Y}_{j}$ and the testing label $Y_{j}$, the estimate error for the parameter $Y_{j}$ can be calculated as,
\begin{equation}
	\label{eq:Err}
	\hat{E}_{j} = ||\hat{Y}_{j} - Y_{j}||./||Y_{j}||
\end{equation} 
This method helps us to assess the change in parameter estimate over the testing/ground-truth values.

%

\section{Experiments and Results} \label{sec:Results}

\subsection{Data} \label{Data}
We perform 4 sets of experiments to assess our proposed method: (1) training and testing on simulated spectra (\textbf{Synthetic - Synthetic (Spectra)}), (2) training and testing on human in-vivo spectral data from different subjects but having the same acquisition protocol (\textbf{Real (Spectra) - Real (Spectra)}), (3) training and testing on human in-vivo spectral data from different subjects with different acquisition protocol (\textbf{Real (Spectra) - Real (MRS Images)}) and (4) using the simulated spectra model to test on MRS images (\textbf{Synthetic (Spectra) - Real (MRS Images)}).

\textbf{Synthetic (Spectra).} \label{Synthetic}
A metabolite basis set was generated by using the data provided by the ISMRM MRS Fitting Challenge 2016. These were then used to simulate over 1 million spectra. In order to ensure that the simulated spectra was as close as possible to human in-vivo spectra, we incorporate the following features: variations in NAA, Cho, Cr concentrations, macro-molecular baseline, lipids, t2 values (for changes in linewidth) and signal-to-noise ratio (SNR) to account for changes in spectral quality. As a preliminary case study, we only simulate the major metabolites (NAA, Cho and Cr) as these are easily detected by the LCModel and would, therefore, help us to evaluate the outcome of our approach and allow a suitable comparison with the LCModel.
  A set of over 10,000 independent test spectra were also simulated with varying combinations of the aforementioned features. For both the training and testing sets, we used the basis-set metabolite concentration values as our ground-truth.
  
\textbf{Real (Spectra).} \label{Real}
To evaluate our method on in-vivo data, we utilize LCModel-fitted single-voxel spectroscopy (SVS) data from 287 independent human subjects. The data was obtained using the same standardized imaging protocol with the following acquisition parameters: TE/TR = 35/2000 ms, spectral width = 2500 Hz, number of points = 1024. We implement a K-fold cross-validation with 10 folds along with the random-forest regression to generate different training and testing sets having spectra from 259 and 28 subjects respectively. The metabolites assessed were: NAA, Cho, mI and Glx.

\textbf{Real (MRS Images).} \label{Real Image}
To further assess our approach, we acquire a standard phase-encoded 2D brain MRSI data of a healthy human volunteer on a 3T scanner using a point-resolved spin-echo localization sequence (PRESS) with voxel size = 10x10x15 mm3, TE/TR=35/1000 ms, spectral width = 2000 Hz, number of points = 400. For testing purposes, we use 96 spectra from the inner-region of the brain which serves as the region of interest.

Due to the differences in acquisition parameters of the training and testing set, both the resulting spectra vary in amplitude and metabolite peak alignment. We perform a pre-processing spectral alignment step where all the test spectra are cropped from 4.3 to 0.2 ppm and interpolated to the same number of points as the training spectra to compensate for differences in acquisition bandwidth. This is followed by normalizing the amplitude of the test spectra using one of the training spectra as reference.

\subsection{Results} \label{sec: Results}
\begin{figure}[ht!]
\floatbox[{\capbeside\thisfloatsetup{capbesideposition={right,top},
capbesidewidth=sidefil}}]{figure}[\FBwidth]  
 {\includegraphics[width=\textwidth, height = 5.5cm, keepaspectratio]{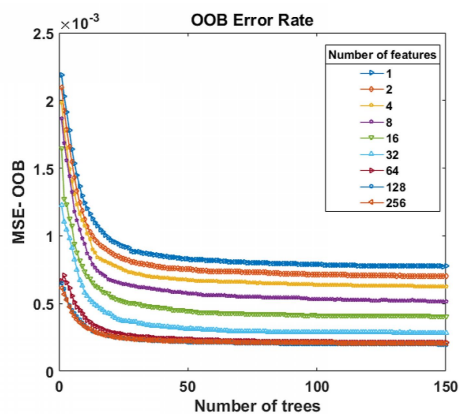}}
 {\caption{Out-Of-Bag (OOB) Error for Simulated Spectra. The experiment is performed for a varying number of features (from 1 to 256 as shown in the legend) and each iteration is assessed for a varying number of trees (as shown in the X-axis). The Y-axis represents the OOB Error rate. The error rate is minimal for more than 64 features and also converges when the number of trees is close to 100.}
 \label{fig:SimOOB}}
\end{figure}

\begin{figure}[ht!]
  \centering
 \includegraphics[width=\textwidth, height = 4cm, keepaspectratio]{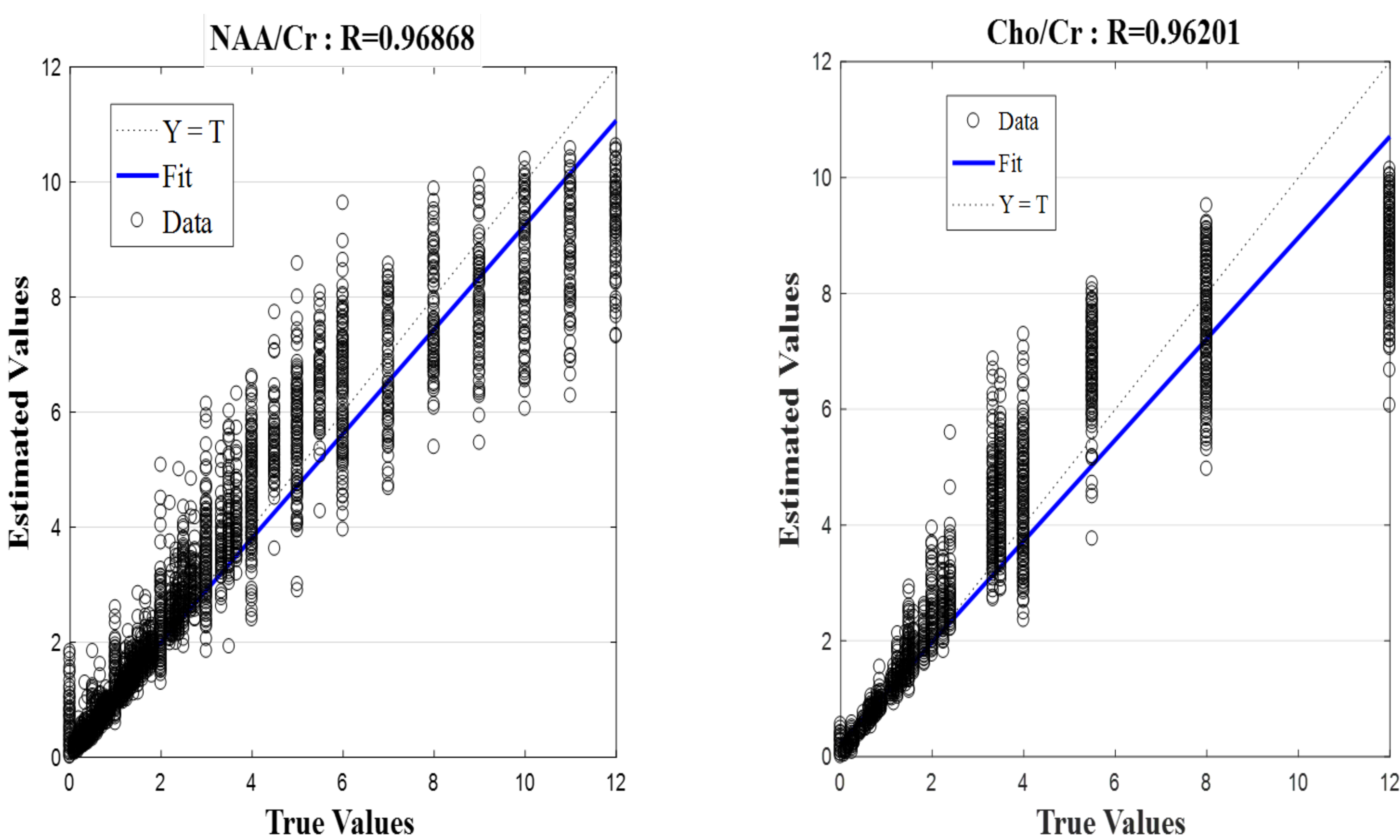}
 \caption{Regression Scores for the following parameters (from left to right): NAA/Cr concentration estimate and Cho/Cr concentration estimate. The X-axis represents the true values of the parameter while the y-axis represents the estimated values. Both sets of values are plotted using linear regression.
 \label{fig:SimRegErr}}
\end{figure}

\begin{figure}[ht!]
  \centering
 \includegraphics[width=\textwidth, height = 4cm, keepaspectratio]{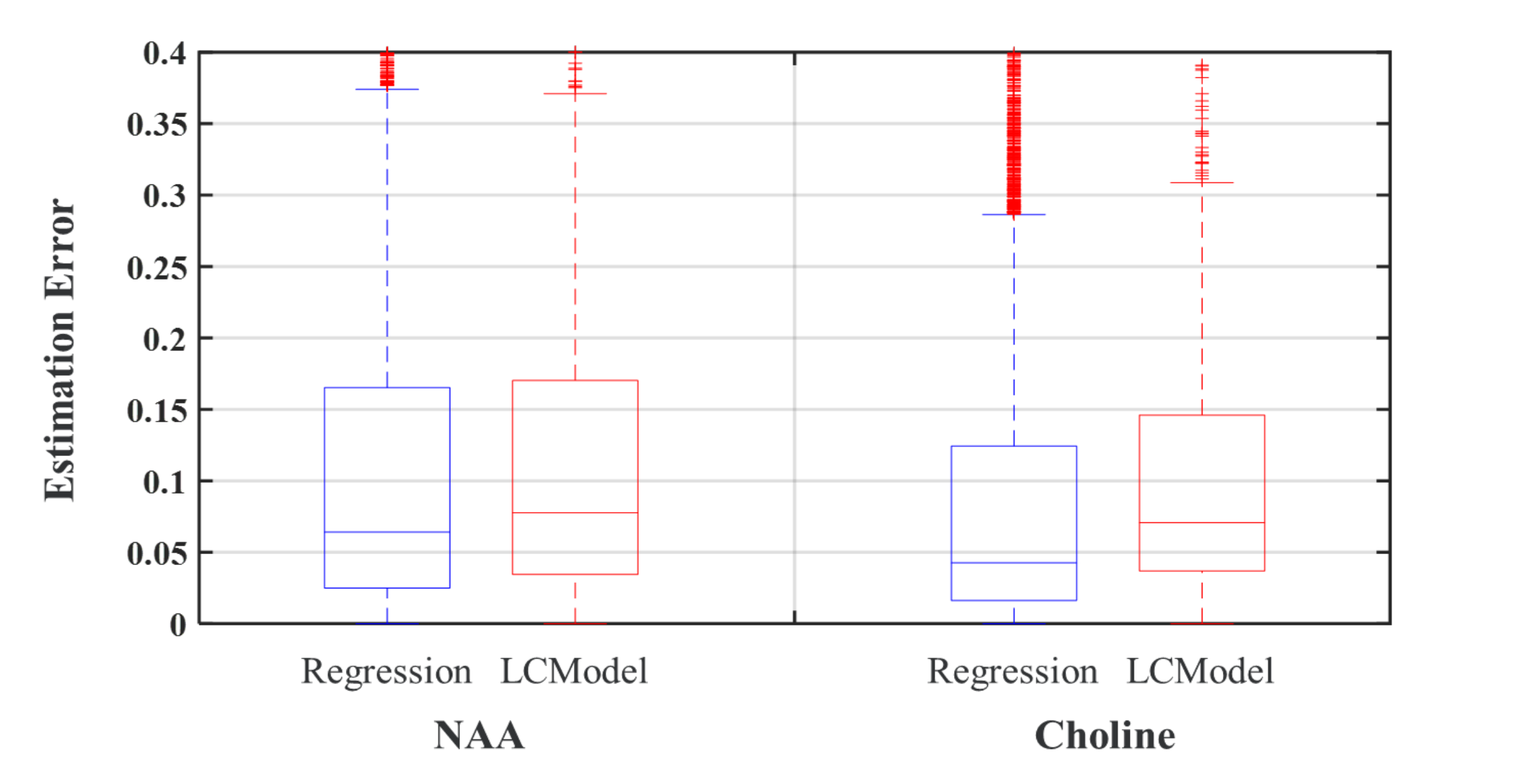}
 \caption{\textbf{Synthetic-Synthetic (Spectra)}: Estimation error for different metabolite concentration ratios in a given test-set. Whiskers span the [min max] values. Median error values are represented by the red line and are as follows: \textbf{NAA/Cr} Regression = 0.064, LCModel = 0.077,  \textbf{Cho/Cr} Regression = 0.043, LCModel = 0.070.
 \label{fig:regLCM}}
\end{figure}

\textbf{Synthetic - Synthetic (Spectra).}
We perform an initial experiment to determine the out-of-bag (OOB) error using different number of trees and features on a set of 20,000 simulated train and test spectra. Based on the results shown in Fig. \ref{fig:SimOOB}, we proceed with the parameter estimation experiment by identifying the appropriate number of trees and features required to achieve convergence of the OOB error. For the regression error estimates, we use metabolite concentration ratios with respect to Cr (used as a standard assessment method in MRS as a means for calibration). We obtain R scores of 0.968 and 0.962 for NAA/Cr and Cho/Cr values respectively. The corresponding figures representing the linear regression are shown in Fig.\ref{fig:SimRegErr} and the error plots in comparison with the LCModel are shown in Fig.\ref{fig:regLCM}.

\begin{figure}[t]
 \begin{floatrow}
  \centering 
  \ffigbox[\textwidth][]{
 \includegraphics[width=0.5\textwidth, height  = 5.5cm, keepaspectratio]{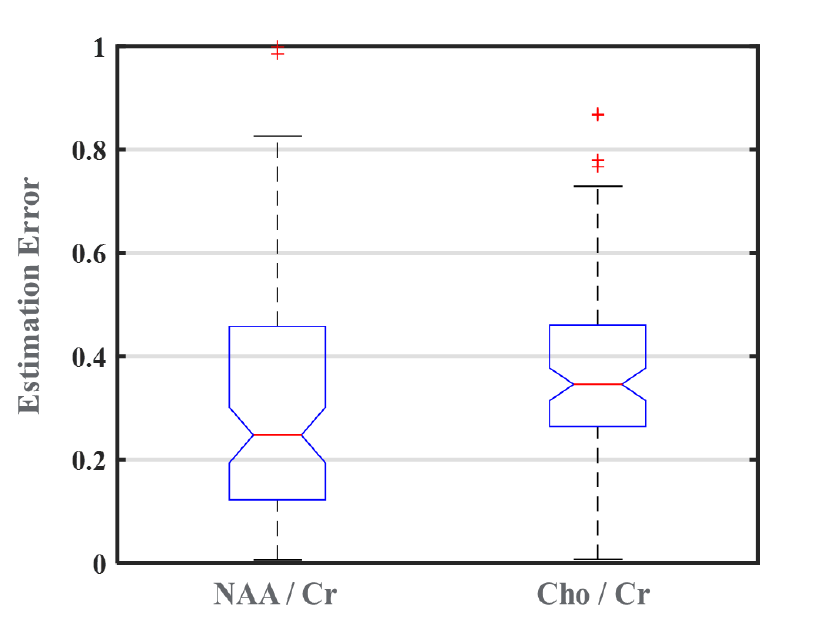}
 \hspace{-0.2cm}
 \includegraphics[width=0.5\textwidth, height = 5.5cm, keepaspectratio]{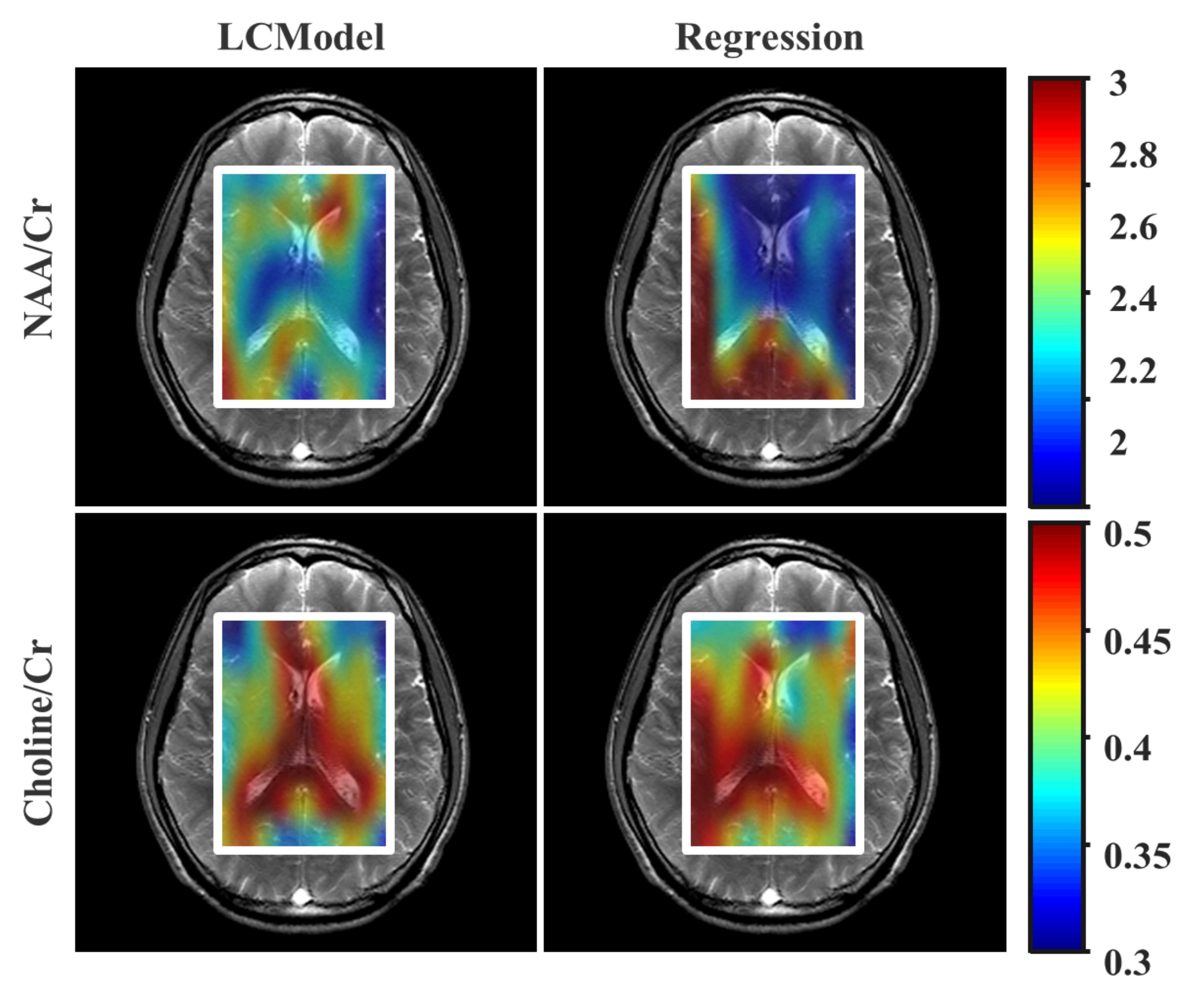}
 }
 {\caption{\textbf{Left:} Synthetic (Spectra)-Real (MRS Images): Estimation error for different metabolite concentration ratios for the same test dataset. Whiskers span the [min max] values. Median error values are represented by the red line and are as follows: NAA/Cr = 0.024, Cho/Cr = 0.034. \textbf{Right:} NAA/Cr and Cho/Cr concentration distribution estimates from random forest regression and non-linear model fit. } 
 \label{fig:synRealErr}}
 \end{floatrow}
\end{figure}

\textbf{Real (Spectra) - Real (Spectra).}
For the SVS dataset, we use the LCModel concentration ratio estimates as the ground-truth. Table.\ref{tab:specerr} indicates the mean metabolite concentration estimate error across the 10-folds of the cross-validation process using the random forest regression method. Median error for the NAA/Cr estimate is 0.068, 0.072 for the Cho/Cr estimate, 0.093 for the mI/Cr estimate and 0.070 for the Glx/Cr estimate compared to the corresponding LCModel estimates. The difference in error estimates is small and shows a similarity in assessment between our proposed method and the LCModel. Moreover, the low-concentration metabolites such as mI and Glx usually display a fitting error with the LCModel and the estimation error for these metabolite ratio concentrations is lower indicating that our model works well for these metabolites as well.

\begin{table}[ht!]
  \begin{floatrow}
    \scriptsize{ \input{realerr-scores}
    }{\caption{Concentration-ratio estimate errors using random forest regression. Results are for the experiments \textbf{Real(spectra)-Real(spectra)} and \textbf{Real(spectra)-Real(Images)}. The errors are calculated over the respective LCModel estimates as per the formula given in Eq.\ref{eq:Err}. The major metabolites (\textbf{NAA} and \textbf{Cr}) show a low error while the smaller concentration metabolites (\textbf{mI} and \textbf{Glx}) show a slightly higher error.}
    \label{tab:specerr}}
\end{floatrow}
\end{table}

\textbf{Synthetic (Spectra) - Real (Images).}
We test our synthetic spectra training model on the 2D MRSI data and the results are shown in the boxplot in Fig. \ref{fig:synRealErr} along with the resulting concentration distribution from both the regression approach and the non-linear model fit. As our synthetic model is trained for only NAA and Cho ratios, we show the errors for these two only. Median estimate error for NAA/Cr is 0.24 using regression. For Cho/Cr, the estimation error is 0.34. The corresponding concentration values estimated from the LCModel serves as our ground-truth.

\textbf{Real (Spectra) - Real (Images).}
We perform a blind test with 96 2D MRSI spectra against the training model generated using the 287 SVS spectra and the results are shown in Table. \ref{tab:specerr}. Median estimate error for NAA/Cr is 0.1, for Cho/Cr is 0.18, for mI/Cr is 0.217 and for Glx/Cr is 0.13. Although we expect the errors to be higher in the blind test due to difference in the acquisition protocols of the training and testing dataset, the errors appear to be within a reasonable window. As expected, the estimated errors are highest for mI/Cr while Glx/Cr surprisingly has a lower error than Cho/Cr.

The Real Spectra training model provides a marginally better metabolite concentration estimate than the Synthetic spectra model . We attribute this to the presence of arbitrary scanning effects and artifacts in the real spectra model as compared to the synthetic model. For future experiments, this provides the scope for learning on a large synthetic spectral data-set with similar additional arbitrary effects to have a robust classifier for real data (especially in the cases where annotating training data is expensive).

\section{Conclusion} \label{sec:Conclusion}

Machine learning techniques such as Random Forest-based regression provide a new and faster way of metabolite quantification. Our synthetic training model accounts for spectral features such as macro-molecular baseline, lipids, linewidth and SNR variations in combination with different metabolite concentrations. Additional features such as frequency and/or phase-shift effects along with B0 inhomogeneity could be incorporated in the model to improve robustness. For the human in-vivo data, we use training spectra from different subjects and the random-forest regression provides a low amount of estimation error over the LCModel fit even in the presence of arbitrary scanning effects. Training times for the simulated spectra can be considerable (around 5-6 hours) given that we generate over 1 million spectra while it is only a few minutes for the in-vivo spectra. On the other hand, testing and concentration estimation happens in only a few seconds and is considerably faster than the non-linear model fitting. The machine learning approach may be used directly, or indirectly by initializing LCModel fits thereby improving their results in the presence of noise and speeding up convergence. They can also be combined with global decisions about spectral quality predicting whether a spectrum can or cannot be interpreted by the physics model because of the presence of artifacts.
 
 Future work would involve using a more robust approach such as deep-learning based methods to improve the accuracy of parameter estimation. Once a framework has been established, further work can be done on having disease-based training models for parameter estimation to predict disease progression and the corresponding metabolite maps.

\textbf{Acknowledgements.} The research leading to these results has received funding from the European Union's H2020 Framework Programme (H2020-MSCA-ITN-2014) under grant agreement n° 642685 MacSeNet.
\bibliographystyle{splncs03}
\bibliography{library}

\end{document}

%% file: realerr-scores.tex
\begin{tabular}{l|rrrrrr}
\toprule
&{ Naa/Cr } & { Cho/Cr } & { mI/Cr } & { Glx/Cr }\\
\midrule
\textbf{Real-Real (Spectra)} & 0.068 & 0.072 & 0.093 & 0.070 \\
\textbf{Real-Real (Images)} & 0.1 & 0.18 & 0.217 & 0.13 \\
\bottomrule
\end{tabular}

%% file: MICCAI17-forGE.bbl
\begin{thebibliography}{1}
\providecommand{\url}[1]{\texttt{#1}}
\providecommand{\urlprefix}{URL }

\bibitem{MRM:MRM26618}
Pedrosa~de Barros, N., McKinley, R., Wiest, R., Slotboom, J.: Improving
  labeling efficiency in automatic quality control of mrsi data. Magnetic
  Resonance in Medicine pp. n/a--n/a (2017),
  \url{http://dx.doi.org/10.1002/mrm.26618}

\bibitem{Breiman2001}
Breiman, L.: Random forests. Machine Learning  45(1),  5--32 (2001),
  \url{http://dx.doi.org/10.1023/A:1010933404324}

\bibitem{Kelm2012}
Kelm, B.M., Kaster, F.O., Henning, A., Weber, M.A., Bachert, P., Boesiger, P.,
  Hamprecht, F.A., Menze, B.H.: {Using spatial prior knowledge in the spectral
  fitting of MRS images}. NMR in Biomedicine  25(1),  1--13 (2012)

\bibitem{MRM:MRM21519}
Menze, B.H., Kelm, B.M., Weber, M.A., Bachert, P., Hamprecht, F.A.: Mimicking
  the human expert: Pattern recognition for an automated assessment of data
  quality in mr spectroscopic images. Magnetic Resonance in Medicine  59(6),
  1457--1466 (2008), \url{http://dx.doi.org/10.1002/mrm.21519}

\bibitem{Provencher1993}
Provencher, S.W.: {Estimation of metabolite concentrations from localized in
  vivo proton NMR spectra.} Magnetic Resonance in Medicine (6),  672--9

\bibitem{RobinA.deGraaf2013}
{Robin A. de Graaf}: {In Vivo NMR Spectroscopy: Principles and Techniques}.
  John Wiley {\&} Sons, 2013, 2 edn. (2013)

\bibitem{Schulte2006}
Schulte, R.F., Lange, T., Beck, J., Meier, D., Boesiger, P.: {Improved
  two-dimensional J-resolved spectroscopy}. NMR in Biomedicine  19(2),
  264--270 (2006)

\bibitem{VANHAMME199735}
Vanhamme, L., van~den Boogaart, A., Huffel, S.V.: Improved method for accurate
  and efficient quantification of mrs data with use of prior knowledge. Journal
  of Magnetic Resonance  129(1),  35 -- 43 (1997),
  \url{http://www.sciencedirect.com/science/article/pii/S1090780797912441}

\bibitem{wilson2011a}
Wilson, M., Reynolds, G., Kauppinen, R.A., Arvanitis, T.N., Peet, A.C.: A
  constrained least‐squares approach to the automated quantitation of in vivo
  1h magnetic resonance spectroscopy data. Magnetic Resonance in Medicine
  65(1),  1--12 (1 2011), \url{http:https://dx.doi.org/10.1002/mrm.22579}

\end{thebibliography}
